%% file: IEEE-conference-template-062824.tex
\documentclass[conference]{IEEEtran}
\IEEEoverridecommandlockouts
\usepackage{url}
\usepackage{cite}
\usepackage{amsmath,amssymb,amsfonts}
\usepackage{algorithmic}
\usepackage{graphicx}
\usepackage{textcomp}
\usepackage{xcolor}
\usepackage{enumerate}
\usepackage{makecell}
\usepackage{array}  
\newcolumntype{B}{!{\vrule width 2\arrayrulewidth}} 
\newcommand{\bhline}{\Xhline{2\arrayrulewidth}}
\def\BibTeX{{\rm B\kern-.05em{\sc i\kern-.025em b}\kern-.08em
    T\kern-.1667em\lower.7ex\hbox{E}\kern-.125emX}}
\begin{document}

\title{\textbf{GO}: The \textbf{G}reat \textbf{O}utdoors Multimodal Dataset\\
}

\author{
\IEEEauthorblockN{Peng Jiang\textsuperscript{1}}
\IEEEauthorblockA{maskjp@tamu.edu}
\and

\IEEEauthorblockN{Kasi Viswanath\textsuperscript{1}}
\IEEEauthorblockA{kasiv@tamu.edu}
\and

\IEEEauthorblockN{Akhil Nagariya\textsuperscript{1}}
\IEEEauthorblockA{akhil.nagariya@tamu.edu}
\and

\IEEEauthorblockN{George Chustz\textsuperscript{1}}
\IEEEauthorblockA{gchustz@tamu.edu}
\and

\IEEEauthorblockN{Maggie Wigness\textsuperscript{2}}
\IEEEauthorblockA{maggie.b.wigness.civ@army.mil}
\and

\IEEEauthorblockN{Philip Osteen\textsuperscript{2}}
\IEEEauthorblockA{philip.r.osteen.civ@army.mil}
\and

\IEEEauthorblockN{Timothy Overbye\textsuperscript{2}}
\IEEEauthorblockA{Tim.Overbye@gmail.com}
\and

\IEEEauthorblockN{Christian Ellis\textsuperscript{2,3}}
\IEEEauthorblockA{christian.ellis@austin.utexas.edu}
\and

\IEEEauthorblockN{Long Quang\textsuperscript{2}}
\IEEEauthorblockA{long.p.quang.civ@army.mil}
\and

\IEEEauthorblockN{Jia Huang\textsuperscript{1}}
\IEEEauthorblockA{jia.huang@tamu.edu}
\and

\IEEEauthorblockN{Srikanth Saripalli\textsuperscript{1}}
\IEEEauthorblockA{ssaripalli@tamu.edu}

\thanks{\textsuperscript{1} Mike Walker ’66 Department of Mechanical Engineering, Texas A\&M University, College Station, TX, USA.}

\thanks{\textsuperscript{2} DEVCOM Army Research Laboratory, Adelphi, MD, USA.}

\thanks{\textsuperscript{3}University of Texas at Austin Center for Autonomy, Austin, TX, USA.}

}

\maketitle

\begin{abstract}
The \textbf{G}reat \textbf{O}utdoors (GO) dataset is a multi-modal annotated data resource aimed at advancing ground robotics research in unstructured environments.
Existing off-road datasets often lack sensor diversity and exclude vital modalities like thermal and radar that are critical for operation in degraded conditions (e.g., low visibility or adverse weather). To address these gaps, we introduce a large-scale multimodal off-road dataset with six complementary sensor modalities, along with semantic annotations and GPS traces, to support tasks such as semantic segmentation, object detection, and SLAM.
The diverse environmental conditions represented in the dataset present significant real-world challenges, which provide opportunities to develop more robust solutions to support the continued advancement of field robotics, autonomous exploration, and perception systems in natural environments.
The dataset can be downloaded at: \url{https://www.unmannedlab.org/the-great-outdoors-dataset/}
\end{abstract}

\input{sections/introduction}
\input{sections/related_work}
\input{sections/dataset_description}
\input{sections/open_questions}
\input{sections/conclusion}
\bibliographystyle{IEEEtran}
\bibliography{IEEEabrv, bib/reference}

\end{document}

%% file: sections/introduction.tex
\section{Introduction}
Robotics research in unstructured environments has experienced rapid growth due to the increasing demand for autonomous systems capable of  perceiving and navigating in complex terrains. Unlike structured urban environments, natural terrains pose unique challenges, including unpredictable obstacles, varying surface types, dense vegetation, uneven topography, and rapidly changing environmental conditions such as lighting and weather. Effective perception and navigation in these settings require a diverse set of sensory inputs and comprehensive training data that represent these complexities.

Traditional approaches in robotics research have focused extensively on urban and structured environments, where predictable roads, clear boundaries, and organized infrastructure simplify  perception and navigation tasks. However, in unstructured environments, such as forests, mountainous terrains, and rural off-road areas, robots must handle a multitude of dynamic variables that require advanced perception capabilities. As the applications for robotics expand into areas such as agriculture, environmental monitoring, search and rescue, and exploration, the need for datasets that accurately reflect these environments has become increasingly urgent.
\input{figures/trails}
Existing datasets used for robotics research are often limited in terms of the diversity of environments, the variety of sensors used, or their applicability to real-world conditions. Many well-known datasets, such as the KITTI Vision Benchmark Suite\cite{geigerVisionMeetsRobotics2013a}, Cityscapes\cite{cordtsCityscapesDatasetSemantic2016} and nuScenes\cite{caesarNuScenesMultimodalDataset2020a}, focus primarily on urban driving and do not offer the environmental diversity needed for field robotics. Existing off-road datasets, such as RELLIS-3D\cite{jiangRELLIS3DDatasetData2022} and GOOSE-Ex\cite{hagmannsExcavatingWildGOOSEEx2024}, have made progress in capturing unstructured environments, but they still have limitations in terms of the sensing modalities included. For instance, many of these datasets do not incorporate thermal or radar sensors which are crucial for perception under challenging conditions like low visibility or adverse weather.
\input{figures/data_showcase2}

To address these limitations, we propose the Great Outdoor (GO) dataset, a diverse and comprehensive multi-modal dataset designed specifically for unstructured environments. By capturing varied terrains such as dense forests, rocky trails, open fields, and bodies of water, as well as utilizing multiple sensor modalities, including camera, stereo, thermal, LiDAR, and radar, the GO dataset provides a rich foundation for advancing the state of the art in field robotics. The inclusion of thermal and radar data which are less commonly found in existing datasets offers a significant advantage for perceiving and navigating in challenging conditions such as low visibility or adverse weather. The GO dataset's design reflects the realities of off-road and natural environments, making it an essential tool for developing resilient and adaptable autonomous systems.

In summary, the GO dataset aims to bridge the gap between current datasets by offering a more diverse, multi-modal, and well-annotated resource for robotics research in natural, unstructured environments. The key contributions of the GO dataset include:
\begin{itemize}
    \item A comprehensive sensor suite that integrates multiple modalities: camera, stereo, thermal, LiDAR, and radar.
    \item Inclusion of thermal and radar data, which are not commonly found in other datasets, to enhance perception robustness for off-road applications.
    \item High quality semantic annotations for images, supporting tasks such as semantic segmentation, object detection, and SLAM.
    \item Coverage of diverse terrains, including those found in forested areas, rocky trails, and other natural environments.
\end{itemize}

%% file: figures/trails.tex
\begin{figure}[]
  \centering
  \includegraphics[width=0.9\columnwidth]{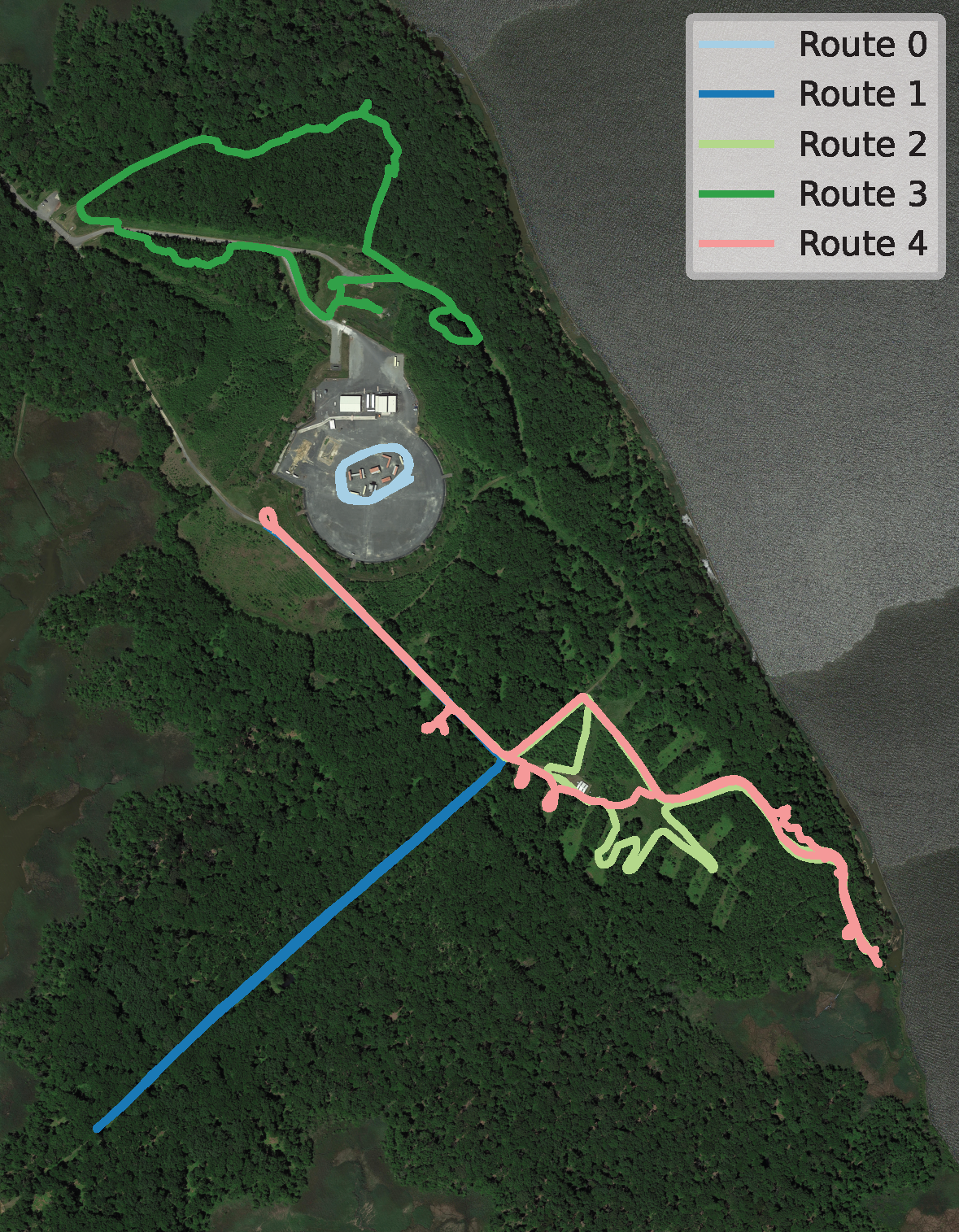}
  \caption{The GO dataset comprises five routes, covering a cumulative distance of 10.26 km and a total duration of 98.60 minutes.}
  \label{fig:trails}
\end{figure}

%% file: figures/data_showcase2.tex
\begin{figure*}[]
\centering
\begin{tabular}{c@{\hspace{1mm}}c@{\hspace{1mm}}c@{\hspace{1mm}}c}
\setlength{\tabcolsep}{3pt}
\includegraphics[width=1.7in,height=1in]{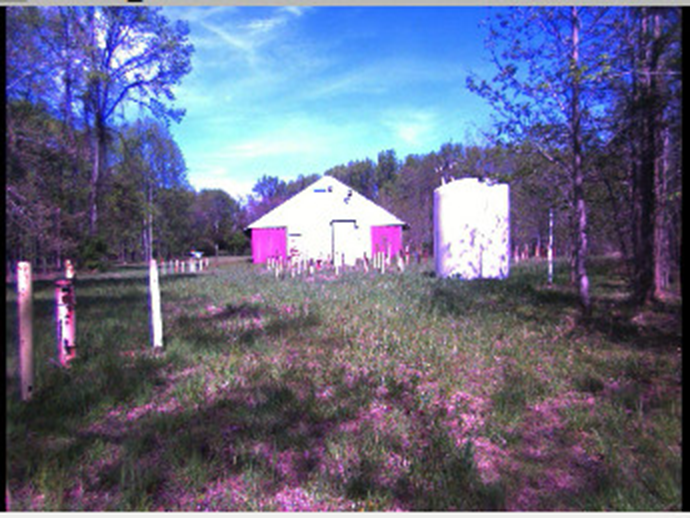}&
\includegraphics[width=1.7in,height=1in]{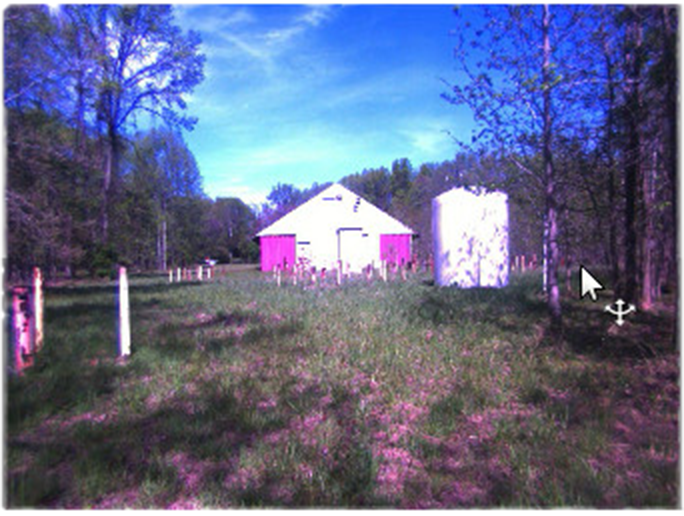}&
\includegraphics[width=1.7in,height=1in]{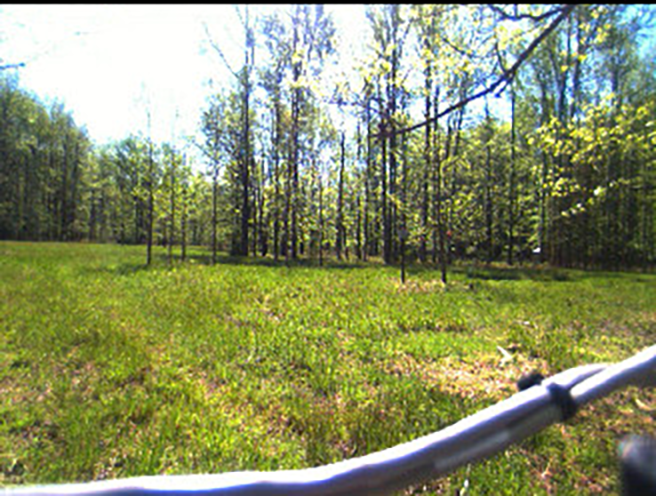}&
\includegraphics[width=1.7in,height=1in]{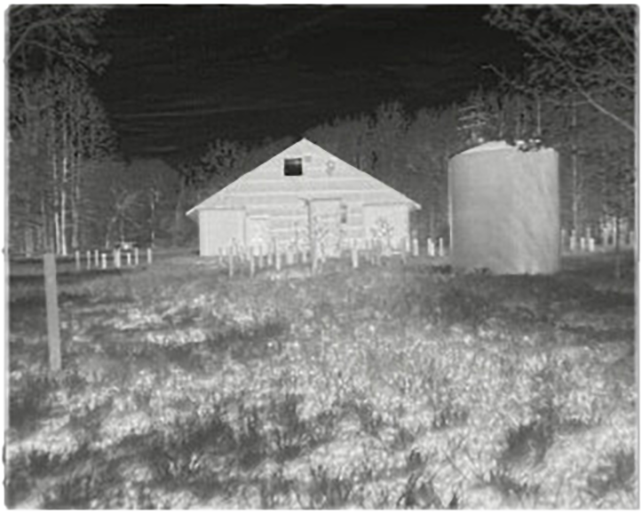}\\
(a) Stereo-Left Camera      & (b) Stereo-Right Camera & (c) Rear Camera & (e) Thermal Camera\\
\multicolumn{4}{c}{\includegraphics[width=7in,height=2in]{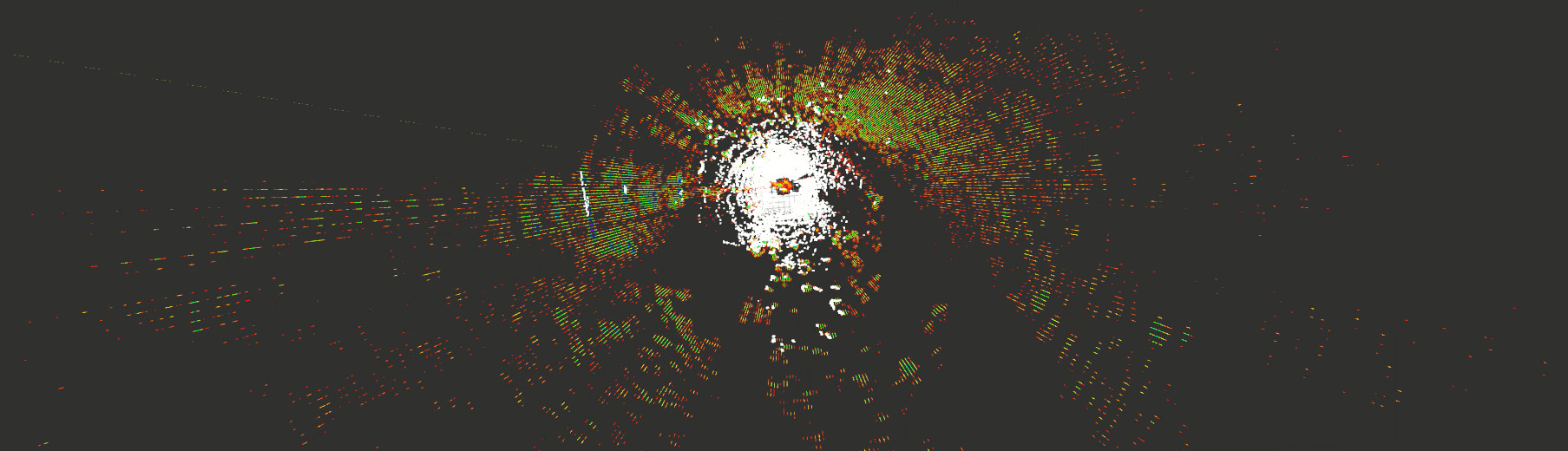}} \\
\multicolumn{4}{c}{(e) Lidar data (white) and Threshold filtered Radar data (color)}
\end{tabular}
\caption{Example of the raw perception sensor data from the GO Dataset. The figure shows (a) the left stereo camera view, (b) the right stereo camera view, (c) the rear camera view, (d) thermal camera imagery, and (e) a combined representation of LiDAR (white) and threshold-filtered radar data (color), where color encodes radar intensity.}
\label{fig:data_showcase}
\end{figure*}

%% file: sections/related_work.tex
\input{tables/datasets}
\section{Related Work}
Several datasets have been developed to support perception systems in off-road robotics. Multi-modal sensing is critical for robust perception in off-road robotics due to the complex and variable nature of the terrains. As one of the early efforts, RELLIS-3D \cite{jiangRELLIS3DDatasetData2022} focuses on unstructured environments, particularly rural and natural settings. It includes LiDAR and RGB data along with their corresponding semantic annotations, providing essential depth and visual information for off-road robotics. The GOOSE-Ex dataset \cite{hagmannsExcavatingWildGOOSEEx2024} offers multi-platform data, including visual and LiDAR sensors, for semantic segmentation in challenging off-road conditions. TartanDrive 2.0 \cite{sivaprakasamTartanDriveMoreModalities2024} is an enhanced version of the original TartanDrive, incorporating an expanded sensor suite that includes multiple LiDAR units alongside cameras and inertial sensors. It covers diverse off-road terrains, providing a comprehensive dataset for self-supervised learning in off-road conditions.

However, present autonomy often depends on exteroceptive sensors like LiDAR and camera. These sensors are highly susceptible to interference from large particles like dust, rain, and snow due to their limited wavelength detection range (camera 400-700 nm and LiDAR 750 nm to 1.5 µm). They can be unreliable in various conditions such as low light, moonlight, darkness, dust (1-400 µm wavelength), smoke (0.1-2.5 µm), fog (10-50 µm), snow, and rain (2 mm). Therefore, datasets that explore other sensors, such as radar and Near-Infrared(NIR) cameras, are essential for off-road environments. OORD \cite{gaddOORDOxfordOffroad2024a} provides radar data for place recognition in rugged, off-road environments, contributing significantly to radar research in unstructured settings. The FoMo dataset \cite{boxanFoMoProposalMultiSeason2024a} provides LiDAR and radar data for navigation in boreal forests, emphasizing sensor fusion in complex natural environments. The TAS-NIR dataset \cite{mortimerTASNIRVIS+NIRDataset2022a} consists of paired visible and near-infrared (VIS+NIR) images, with fine-grained semantic segmentation of vegetation and ground surfaces in unstructured outdoor environments. The NIR modality helps improve segmentation accuracy for vegetation and other natural elements, making it valuable for applications in autonomous off-road driving. 

For a comprehensive overview of existing datasets, we refer the reader to the survey by Mortimer and Maehlisch \cite{mortimerSurveyDatasetsPerception2024}, which highlighted key limitations in current datasets. Despite progress, there is still a lack of datasets that encompass all sensing modalities while providing semantic segmentation labels (see Table. \ref{table:datasets}). The GO dataset aims to address these gaps by incorporating all the modalities and offering detailed semantic annotations.

%% file: tables/datasets.tex
\begin{table}
\caption{Comparison of Several Existing Off-road Multi-modal Datasets}
\centering
\begin{tabular}{c@{\hspace{1mm}}c@{\hspace{1mm}}c@{\hspace{1mm}}c@{\hspace{1mm}}c@{\hspace{1mm}}c@{\hspace{1mm}}c@{\hspace{1mm}}c@{\hspace{1mm}}c}
\bhline
Dataset& Camera & Stereo & LIDAR & NIR & Radar & IMU & GPS & Labels \\ \bhline
RELLIS-3D\cite{jiangRELLIS3DDatasetData2022}  &$\bullet$ &$\bullet$  &$\bullet$&$\circ$        &$\circ$  &$\bullet$&$\bullet$&$\bullet$\\
TartanDrive 2.0\cite{sivaprakasamTartanDriveMoreModalities2024}& $\bullet$ &$\bullet$  &$\bullet$&$\bullet$        &$\circ$  &$\bullet$&$\bullet$&$\circ$\\
OORD  \cite{gaddOORDOxfordOffroad2024a}    &$\bullet$ &$\circ$  &$\circ$&$\circ$        &$\bullet$  &$\bullet$&$\bullet$&$\circ$\\
GOOSE-Ex \cite{hagmannsExcavatingWildGOOSEEx2024} &$\bullet$ &$\bullet$  &$\bullet$&$\bullet$        &$\circ$  &$\bullet$&$\bullet$&$\bullet$\\
FOMO \cite{boxanFoMoProposalMultiSeason2024a}      &$\bullet$ &$\bullet$  &$\bullet$&$\circ$        &$\bullet$  &$\bullet$&$\bullet$&$\circ$\\ 
TAS-NIR \cite{mortimerTASNIRVIS+NIRDataset2022a}      &$\bullet$ &$\circ$  &$\circ$& $\bullet$& $\circ$ & $\circ$& $\circ$& $\bullet$\\ 
\hline
GO        &$\bullet$ &$\bullet$  &$\bullet$&$\bullet$        &$\bullet$  &$\bullet$&$\bullet$&$\bullet$\\ \bhline
\end{tabular}
\label{table:datasets}
\end{table}

%% file: sections/dataset_description.tex
\section{Dataset Descriptions}
\subsection{Sensor Setup}
We used a Clearpath Warthog mobile robot as our platform (see Fig. \ref{fig:sensor}) to gather data (see Fig. \ref{fig:data_showcase}). The onboard sensor suite includes:
\begin{enumerate}
  \item 1 \(\times\) Ouster OS1 LiDAR: 64 Channels, 2048 horizontal resolution, 10 Hz, 45\(^{\circ}\) vertical field of view;
  \item 1 \(\times\) RGB Camera: FLIR Blackfly S, Rear-mounted, running at 30 Hz; 
  \item 1 \(\times\) Stereo Camera: FLIR Blackfly S, Front-facing stereo configuration, 30 Hz;
  \item 1 \(\times\) Thermal Camera: FLIR Boson 640, operating at 60 Hz;
  \item 1 \(\times\) Inertial Navigation System (IMU/GPS): MicroStrain 3DM-GX5-AHRS, provides 200 Hz IMU data;
  \item 2D mmWave Radar: The Navtech CTS350-X functions at 4 Hz, offering $360^{\circ}$ azimuth coverage with a $0.9^{\circ}$ sampling interval, and has a range capacity of 270 meters along with 400 input rotations per cycle;
  \item RTK GPS: The Sparkfun RTK Facet provides a maximum accuracy of 1.4 cm and operates at a frequency of 4 Hz, utilizing a subscription-based NTRIP service for RTCM corrections.
\end{enumerate}
\input{figures/sensors}
\subsection{Sensor Calibration and Synchronization}
To ensure accurate sensor synchronization across the computer and sensor network, we utilize Precision Time Protocol (PTP) for radar, LiDAR, and monocular and stereo cameras. A high-precision clock serves as the authoritative time source that synchronizes sensors and computers via PTP. The thermal camera and GNSS/IMU rely on ROS timestamps for synchronization.

For calibration, we employ the MSG-Cal method \cite{owensMSGcalMultisensorGraphbased2015b}, using a planar board to align the LiDAR and RGB/Thermal camera sensors, as shown in Fig. \ref{fig:calibration}. The radar's transformation with respect to other sensors is calibrated using a 3D model to align the radar data with the 3D scanned point cloud accurately.
\input{figures/calibration}
\subsection{Scene Information}
Data acquisition was conducted over a two-day period in May at a robotics research facility with 200 acres of natural terrain representing diverse regions of forested terrain, unpaved trails, and gravel areas. 
The dataset comprises a teleoperated collection of five distinct routes (as seen in Fig.~\ref{fig:trails}), collectively spanning 10.26 km of unstructured terrain with a combined total operation time of 98.60 minutes. Route descriptions are as follows:  
\begin{itemize}
    \item \textit{Route 0} serves as a test route where the robot travels at high speed in a minimally featured gravel area, covering 0.58 km in 3.43 minutes, with a peak speed of 4.00 m/s. 
    \item \textit{Route 1} features a loop on a gravel roadway, spanning 2.35 km in 13.02 minutes and an average speed of 3.15 m/s. 
    \item \textit{Route 2} was recorded on a trail with multiple instances of the robot deviating into off-road areas before returning to the main path, achieving an average speed of 2.80 m/s. 
    \item \textit{Route 3} is the most intricate route among the five, integrating both gravel trails and forest regions, resulting in an average speed of just 0.98 m/s due to difficult terrain. 
    \item \textit{Route 4} follows the same trail as \textit{Route 2} but on a different day, and deviates into different off-road areas as shown in Fig. \ref{fig:trails}, resulting in an average speed of 1.59 m/s.
\end{itemize}
\input{figures/data_embeding}
\input{tables/tracks}
A summary of the time, distance, velocity, and environment type for each route is presented in Table \ref{table:tracks}.

Representative scenes from the routes are visualized in Fig.~\ref{fig:go_embeding}, where each image is encoded into a 512-dimensional CLIP embedding~\cite{radfordLearningTransferableVisual2021} and then projected into a 2D space using t-SNE~\cite{caiTheoreticalFoundationsTSNE2022}. The embeddings are grouped into 10 clusters via K-Means, and representative images closest to each cluster centroid are displayed alongside the embedding map.

\input{tables/merged_iou}
\subsection{Semantic Annotations}
\subsubsection{Ontology}The GO dataset provides detailed pixel-wise semantic annotations (see Fig.~\ref{fig:labels}) to support enhanced autonomous off-road navigation. By integrating the ontological frameworks of the RELLIS-3D~\cite{jiangRELLIS3DDatasetData2022} and RUGD~\cite{wignessRUGDDatasetAutonomous2019} datasets, we construct a comprehensive ontology of terrain and object categories tailored for the GO dataset. In total, the dataset includes 22 distinct classes, covering categories including trees, grass, dirt, sky, gravel, bush, mulch, water, poles, fences, persons, buildings, objects, vehicles, barriers, mud, concrete, puddles, rubble, asphalt, and a void class.

\subsubsection{Annotation Process}The annotation process was semi-automated: we selected keyframes and used the Segment Anything Model (SAM) \cite{kirillovSegmentAnything2023} and OffSeg model \cite{viswanathOffsegSemanticSegmentation2021b} to provide initial segmentation labels. These initial labels were then refined by human annotators who manually assigned semantic labels, adjusted boundaries, and corrected errors. We also provide labels to thermal image data by utilizing the calibration between RGB camera and thermal camera. A total of 1,800 image frames were annotated. 
\input{figures/labels}

\subsubsection{Class Distribution}The semantic class distribution for the GO dataset can be seen in Fig. \ref{fig:go_stat} and reveals that vegetation-related classes, such as trees and grass, are highly represented, while man-made structures like fences, vehicle, and building are less frequent. This imbalance highlights the emphasis on natural terrains, making the GO dataset well-suited for off-road robotics, but also presents challenges for training balanced perception models across all classes. 

\input{figures/go_stat}


\section{Preliminary GO Dataset Benchmarking}
\label{subsec:rgb_thermal_eval}
In this section, we conduct a preliminary evaluation of existing image segmentation approaches to establish a benchmark on the GO dataset. This evaluation specifically focuses on the use of thermal and RGB imagery to (i) assess architecture performance in the off-road domain with a single data modality, i.e., intra-modal evaluation, (ii) assess model generalizability in a cross-modal training/testing setting, i.e., training and testing data modalities differ, and (iii) assess joint RGB-thermal model learning. We choose four representative segmentation models: two CNN-based architectures, UPerNet~\cite{xiaoUnifiedPerceptualParsing2018a} and DeepLabV3+~\cite{chenEncoderDecoderAtrousSeparable2018}, and two transformer-based architectures, Dense Prediction Transformer (DPT)~\cite{ranftlVisionTransformersDense2021b} and SegFormer~\cite{xieSegFormerSimpleEfficient2021a}, which are implemented in \textit{smp library}~\cite{Iakubovskii:2019}. 
These models provide a comprehensive view of segmentation performance trends across different modalities and architectural designs.  

\subsection{Experimental Setup}
The GO dataset is split into \texttt{train}, \texttt{val}, and \texttt{test} sets. Models are trained only with the \texttt{train} set, but for a comprehensive evaluation, performance is reported across all three splits. We perform a time-aware, window-based split to avoid temporal leakage. Labeled frames are first sorted in chronological order and grouped into contiguous windows of fixed length (150 frames by default). The windows are then split contiguously into training, validation, and test sets (70\% / 15\% / 15\%), with a small temporal buffer between splits to prevent overlap. Semantic class coverage is computed per window, and windows are minimally reassigned (while preserving temporal separation) to ensure that all required classes appear in both the training and test sets. 
Details of the three assessments are as follows:
\begin{enumerate}[(i)]
    \item Intra-modal: training and testing modalities are the same (RGB $\rightarrow$ RGB ; Thermal $\rightarrow$ Thermal)
    \item Cross-modal: training and testing modalities are different (RGB $\rightarrow$ Thermal ; Thermal $\rightarrow$ RGB) 
    \item Joint-modal: mulitple modalities make up the training and are tested on a single modality (RGB \& Thermal $\rightarrow$ RGB ; RGB \& Thermal $\rightarrow$ Thermal)
\end{enumerate}
Mean Intersection-over-Union (mIoU) is reported for all settings in Table \ref{tab:rgb_thermal_seg}.  

\subsection{Intra-Modal Performance}
As shown in Table~\ref{tab:rgb_thermal_seg}, all models achieve their highest performance when trained and evaluated on the same modality.
DeepLabV3+, SegFormer, and UPerNet consistently outperform DPT across both RGB and thermal domains, reflecting their stronger architectural adaptation for dense prediction.
DeepLabV3+ performs particularly well on RGB data, benefiting from its atrous spatial pyramid pooling (ASPP) that captures multi-scale context effectively.
SegFormer also delivers strong results, aided by its lightweight hierarchical transformer and fine-grained $4\times4$ patch embeddings, which better preserve spatial detail than DPT’s $14\times14$ tokens.
Among all models, UPerNet achieves the best overall intra-modal results, with 0.78 mIoU on RGB and 0.66 on thermal test sets, confirming the effectiveness of its multi-level feature fusion and semantic pyramid design for both modalities. As shown in the results, performance on thermal-only training remains noticeably lower than RGB, indicating that current architectures still struggle to fully exploit the weaker texture, lower contrast, and noisier structure of thermal imagery. 
\subsection{Cross-Modal Generalization}
When models are trained on one modality and tested on another, performance declines markedly due to spectral and appearance discrepancies between visual and infrared imagery.
For instance, DPT’s mIoU drops from 0.58 (RGB test) to 0.46 when evaluated on thermal data, while UPerNet similarly decreases from 0.78 to 0.53.
DeepLabV3+ and SegFormer exhibit moderate resilience to this shift, suggesting that both convolutional multi-scale features (in DeepLabV3+) and transformer-based global context modeling (in SegFormer) offer some degree of robustness to cross-modal variation.
These results underline the challenge of modality transfer, as visual-textural cues from RGB do not directly align with intensity-based thermal signatures.

\subsection{Joint-Modality Training}
When trained jointly on both RGB and thermal images, all models exhibit more balanced performance across modalities and substantially improved cross-modal generalization. Although joint training does not surpass the highest intra-modal accuracy achieved by single-modality training (e.g., RGB→RGB or Thermal→Thermal), it significantly reduces the performance drop when evaluating on the opposite modality. For example, DPT’s cross-modal mIoU increases from 0.46/0.37 to 0.56/0.51 under joint supervision, while SegFormer and DeepLabV3+ maintain strong performance on both RGB and thermal test sets.
The largest improvement occurs in UPerNet, which achieves 0.75 mIoU on RGB and 0.68 on thermal—representing the best cross-modally consistent result among all models. This shows that multi-scale fusion not only improves representational robustness but also facilitates feature sharing between complementary sensing modalities.

\subsection{Discussion}
These results highlight three main findings:  
(1) semantic annotations are internally consistent across modalities, enabling effective model learning;  
(2) there exists a clear domain gap between RGB and thermal data that limits zero-shot transfer; and  
(3) multi-modal training effectively mitigates this gap by providing cross-spectral supervision.  
In practice, jointly training on both modalities offers the most reliable performance for off-road perception tasks, where lighting, temperature, and visibility conditions can change dramatically.

%% file: figures/sensors.tex
\begin{figure}[]
  \centering
  \includegraphics[width=0.8\columnwidth]{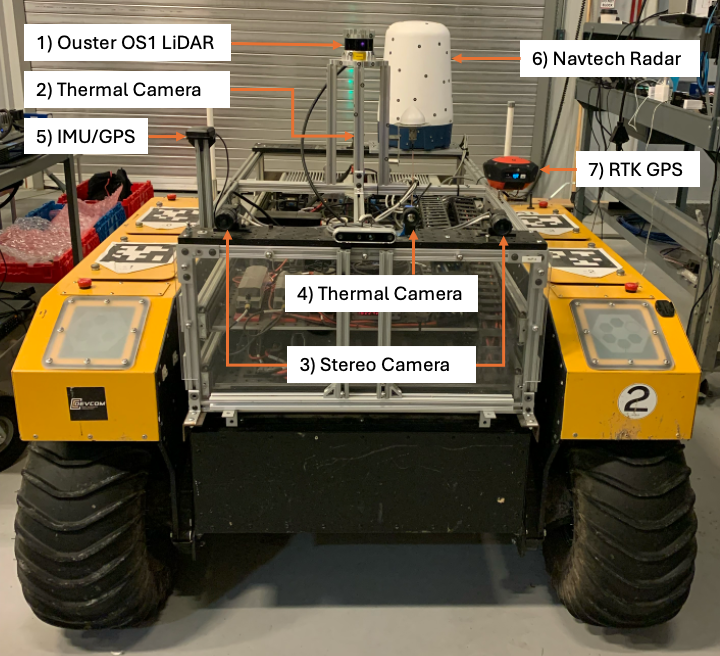}
  \caption{Multimodal sensor suite mounted on the Clearpath Warthog for off-road data collection.}
  \label{fig:sensor}
\end{figure}

%% file: figures/calibration.tex
\begin{figure}[]
  \centering
  \includegraphics[width=0.9\columnwidth]{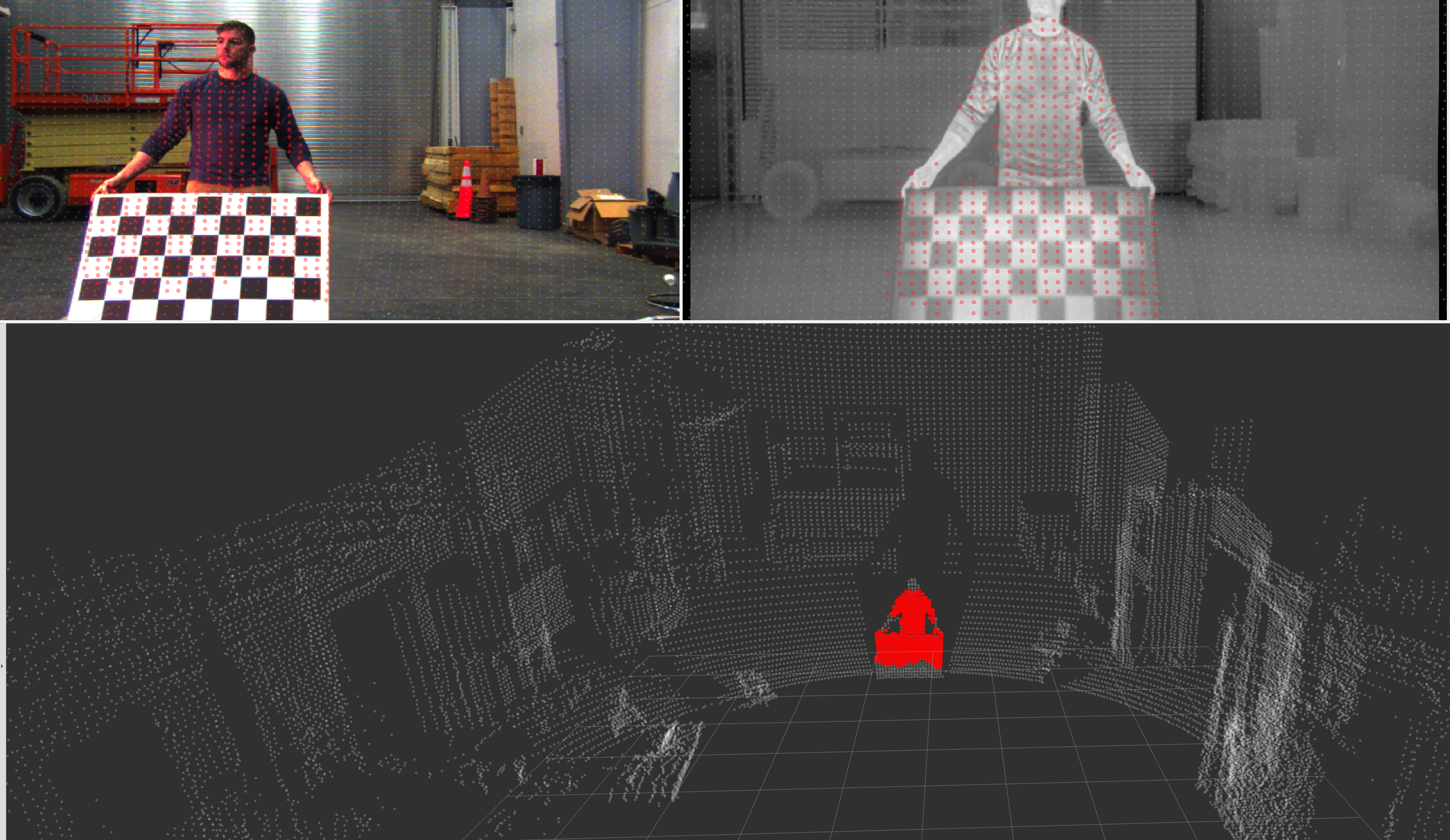}
  \caption{Qualitative visualization of the LiDAR-Camera calibration.}
  \label{fig:calibration}
\end{figure}

%% file: figures/data_embeding.tex
\begin{figure}[]
  \centering
  \includegraphics[width=\columnwidth]{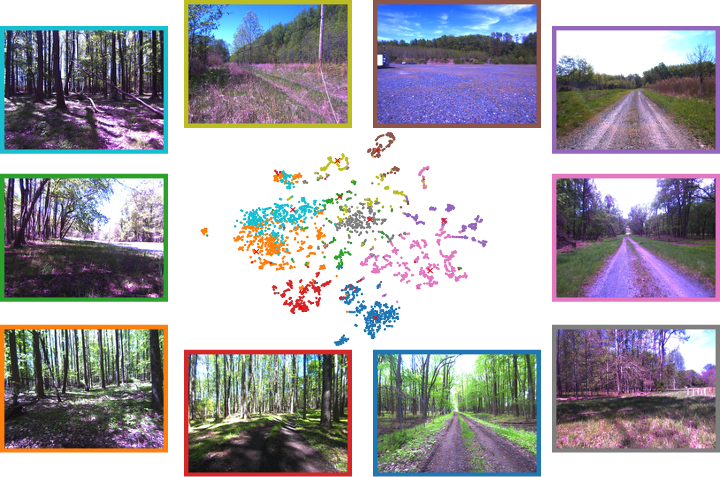}
  \caption{The image data were encoded into 512-dimensional CLIP embeddings, reduced to a 2-D space using t-SNE for visualization, and grouped into 10 clusters via K-Means. Representative images nearest to each cluster center are displayed alongside the embedding map.}
  \label{fig:go_embeding}
\end{figure}

%% file: tables/tracks.tex
\begin{table}[]
\caption{Details of the Five Routes of the GO Dataset}
\begin{tabular}{cBc|c|c|c|c}
\bhline
Route    & $t$(min) & $d$(km) & $v_{max}$(m/s) & $v_{avg}$(m/s) & Environment    \\ \bhline
 0 & 3.43      & 0.58         & 4.00           & 2.91           & gravel    \\ \hline
 1 & 13.02     & 2.35         & 3.98           & 3.15           & trails         \\ \hline
 2 & 17.90     & 2.79         & 4.06           & 2.80           & trails, forest \\ \hline
 3 & 33.40     & 1.82         & 3.18           & 0.98           & trails, forest \\ \hline
 4 & 30.85     & 2.72         & 3.96           & 1.59           & trails, forest \\ \bhline
\end{tabular}
\label{table:tracks}
\end{table}

%% file: tables/merged_iou.tex
\begin{table*}[]
\scriptsize		
\centering
\caption{
Mean Intersection-over-Union (mIoU) of DPT, SegFormer, and UPerNet on RGB and thermal modalities. 
The \textbf{Train/Test} row indicates the training and evaluation modalities, respectively. 
Results are reported for \texttt{train}, \texttt{val}, and \texttt{test} splits. 
Joint RGB \& Thermal training improves generalization across modalities.}
\begin{tabular}{cBccccccBccccccBcccccc}
\bhline
           & \multicolumn{6}{cB}{\textbf{Intra-Modal}}                                                                                                                                                              & \multicolumn{6}{cB}{\textbf{Cross-Modal}}                                                                                                                                                              & \multicolumn{6}{c}{\textbf{Joint-Modal}}                                                                                                                                                               \\ \bhline
Train/Test & \multicolumn{3}{cB}{RGB/RGB}                                                                                 & \multicolumn{3}{cB}{Thermal/Thermal}                                                    & \multicolumn{3}{cB}{RGB/Thermal}                                                                             & \multicolumn{3}{cB}{Thermal/RGB}                                                        & \multicolumn{3}{cB}{RGB \& Thermal/RGB}                                                                      & \multicolumn{3}{c}{RGB \& Thermal/Thermal}                                              \\ \bhline
Split      & \multicolumn{1}{c|}{train}         & \multicolumn{1}{c|}{val}           & \multicolumn{1}{cB}{test}          & \multicolumn{1}{c|}{train}         & \multicolumn{1}{c|}{val}           & \multicolumn{1}{cB}{test}          & \multicolumn{1}{c|}{train}         & \multicolumn{1}{c|}{val}           & \multicolumn{1}{cB}{test}          & \multicolumn{1}{c|}{train}         & \multicolumn{1}{c|}{val}           & \multicolumn{1}{cB}{test}          & \multicolumn{1}{c|}{train}         & \multicolumn{1}{c|}{val}           & \multicolumn{1}{cB}{test}          & \multicolumn{1}{c|}{train}         & \multicolumn{1}{c|}{val}           & test          \\ \hline
DPT        & \multicolumn{1}{c|}{0.69}          & \multicolumn{1}{c|}{0.69}          & \multicolumn{1}{cB}{0.58}          & \multicolumn{1}{c|}{0.62}          & \multicolumn{1}{c|}{0.72}          & \multicolumn{1}{cB}{0.54}          & \multicolumn{1}{c|}{0.47}          & \multicolumn{1}{c|}{0.63}          & \multicolumn{1}{cB}{0.46}          & \multicolumn{1}{c|}{0.40}          & \multicolumn{1}{c|}{0.56}          & \multicolumn{1}{cB}{0.37}          & \multicolumn{1}{c|}{0.65}          & \multicolumn{1}{c|}{0.67}          & \multicolumn{1}{cB}{0.56}          & \multicolumn{1}{c|}{0.56}          & \multicolumn{1}{c|}{0.70}          & 0.51         \\ \hline
Segformer  & \multicolumn{1}{c|}{0.80}          & \multicolumn{1}{c|}{0.83}          & \multicolumn{1}{cB}{0.71}          & \multicolumn{1}{c|}{0.72}          & \multicolumn{1}{c|}{0.75}          & \multicolumn{1}{cB}{0.60}          & \multicolumn{1}{c|}{0.48}          & \multicolumn{1}{c|}{0.72}          & \multicolumn{1}{cB}{0.44}          & \multicolumn{1}{c|}{\textbf{0.63}} & \multicolumn{1}{c|}{\textbf{0.74}} & \multicolumn{1}{cB}{\textbf{0.57}} & \multicolumn{1}{c|}{0.84}          & \multicolumn{1}{c|}{0.83}          & \multicolumn{1}{cB}{0.72}          & \multicolumn{1}{c|}{0.73}          & \multicolumn{1}{c|}{0.76}          & 0.58          \\ \hline
DeepLabV3+ & \multicolumn{1}{c|}{0.88}          & \multicolumn{1}{c|}{0.86}          & \multicolumn{1}{cB}{0.75}          & \multicolumn{1}{c|}{0.80}          & \multicolumn{1}{c|}{0.78}          & \multicolumn{1}{cB}{0.66}          & \multicolumn{1}{c|}{\textbf{0.68}} & \multicolumn{1}{c|}{\textbf{0.74}} & \multicolumn{1}{cB}{\textbf{0.60}} & \multicolumn{1}{c|}{0.33}          & \multicolumn{1}{c|}{0.42}          & \multicolumn{1}{cB}{0.32}         & \multicolumn{1}{c|}{0.67}          & \multicolumn{1}{c|}{0.74}          & \multicolumn{1}{cB}{0.60}          & \multicolumn{1}{c|}{0.80}          & \multicolumn{1}{c|}{0.79}          & 0.67          \\ \hline
UPerNet    & \multicolumn{1}{c|}{\textbf{0.89}} & \multicolumn{1}{c|}{\textbf{0.87}} & \multicolumn{1}{cB}{\textbf{0.78}} & \multicolumn{1}{c|}{\textbf{0.80}} & \multicolumn{1}{c|}{\textbf{0.79}} & \multicolumn{1}{cB}{\textbf{0.66}} & \multicolumn{1}{c|}{0.63}          & \multicolumn{1}{c|}{0.52}          & \multicolumn{1}{cB}{0.53}          & \multicolumn{1}{c|}{0.41}          & \multicolumn{1}{c|}{0.57}          & \multicolumn{1}{cB}{0.40}          & \multicolumn{1}{c|}{\textbf{0.88}} & \multicolumn{1}{c|}{\textbf{0.86}} & \multicolumn{1}{cB}{\textbf{0.75}} & \multicolumn{1}{c|}{\textbf{0.82}} & \multicolumn{1}{c|}{\textbf{0.79}} & \textbf{0.68} \\ \bhline
\end{tabular}

\label{tab:rgb_thermal_seg}
\end{table*}

%% file: figures/labels.tex
\begin{figure}[]
    \centering
\begin{tabular}{c@{\hspace{1mm}}c}
\includegraphics[width=1.7in,height=1in]{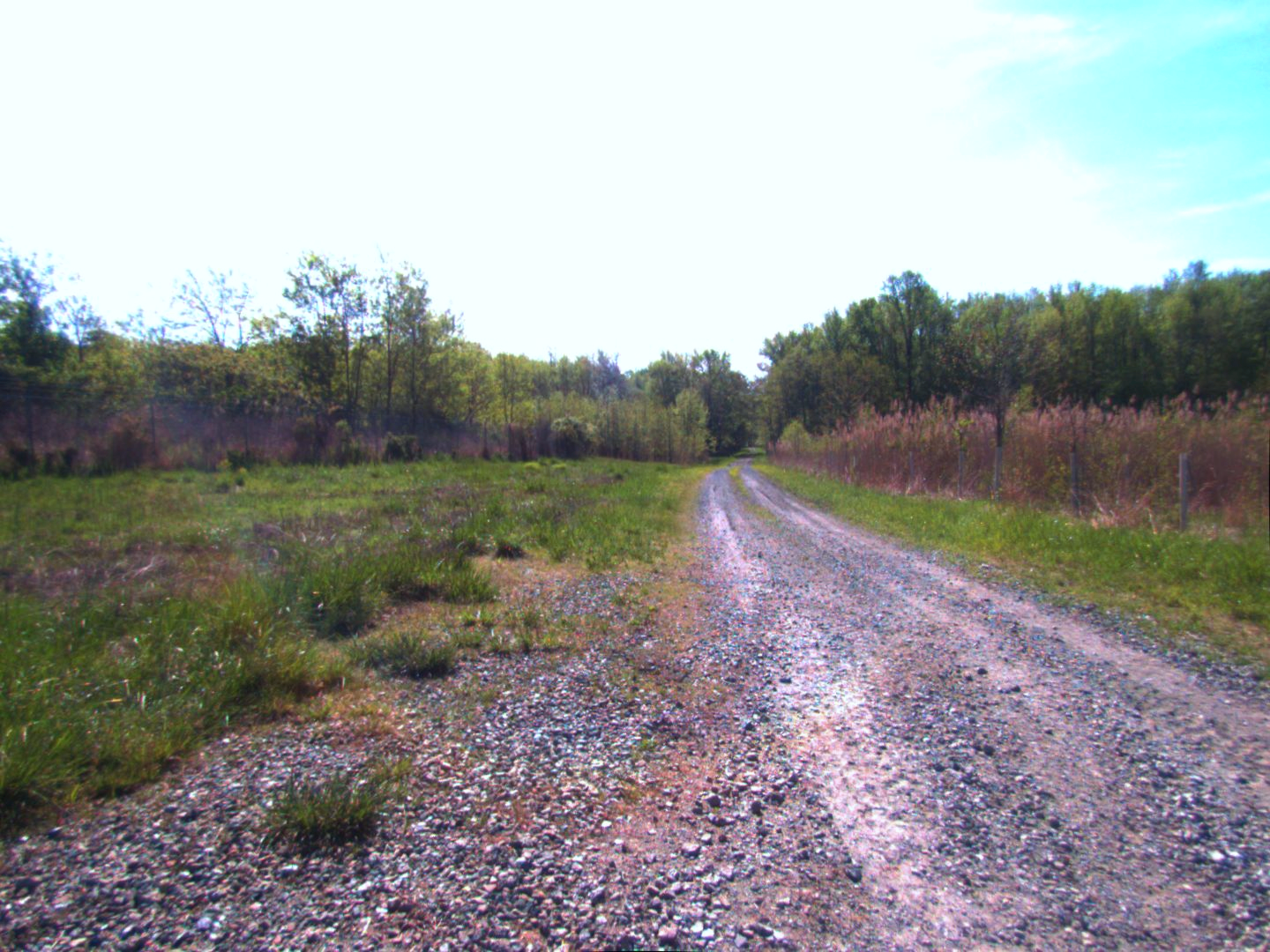}&
\includegraphics[width=1.7in,height=1in]{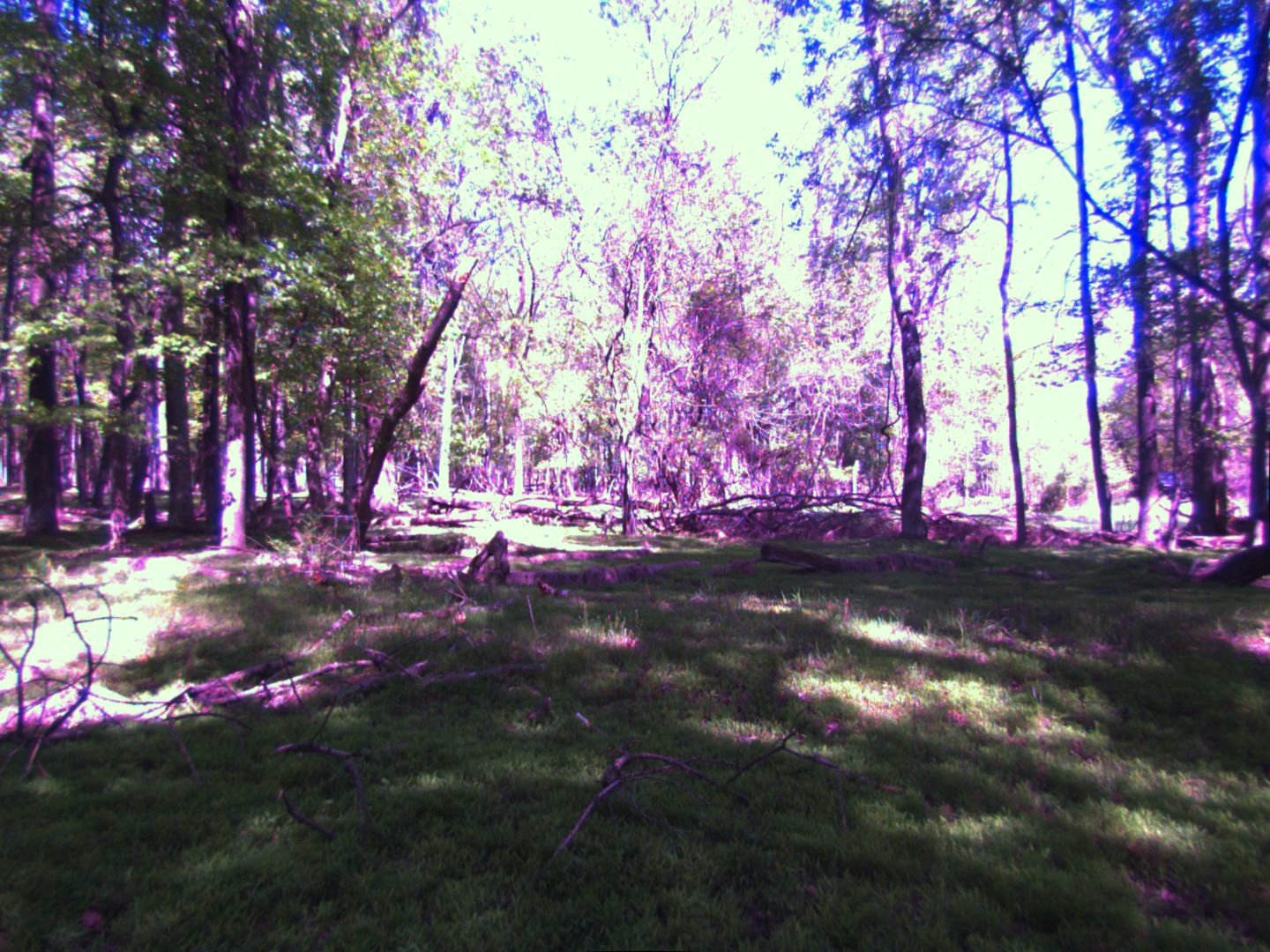}\\
\includegraphics[width=1.7in,height=1in]{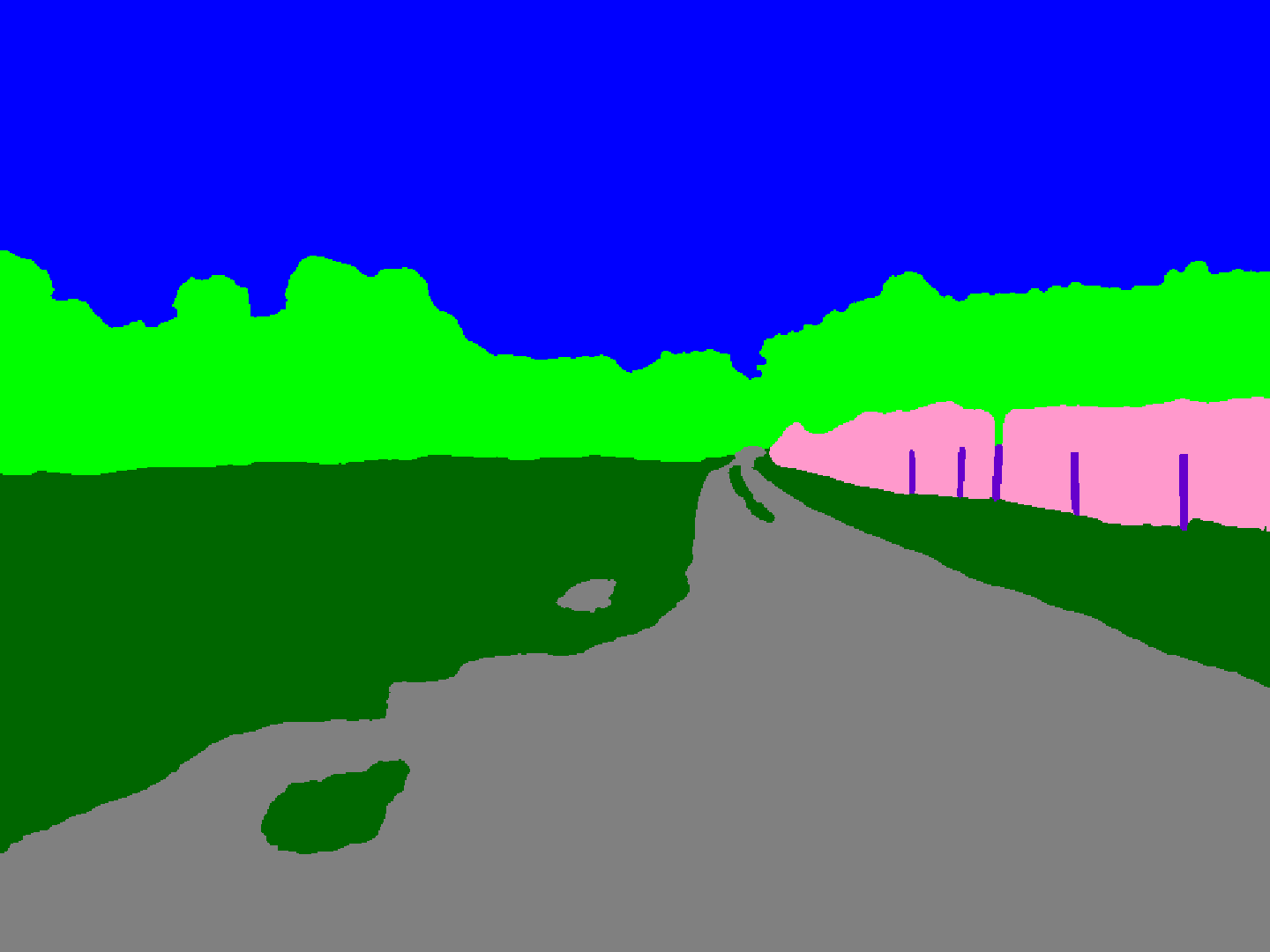}&
\includegraphics[width=1.7in,height=1in]{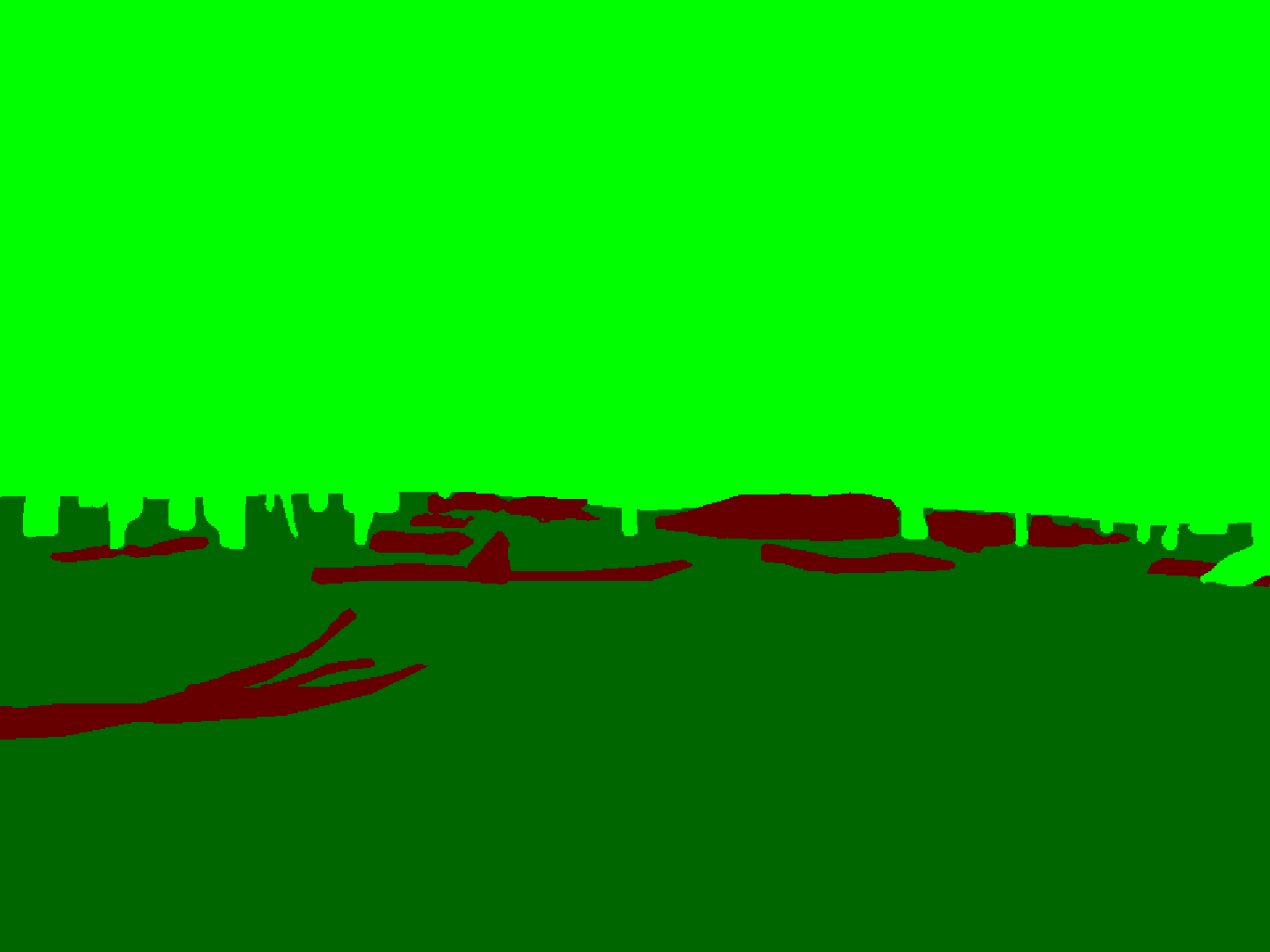}\\
\includegraphics[width=1.7in,height=1in]{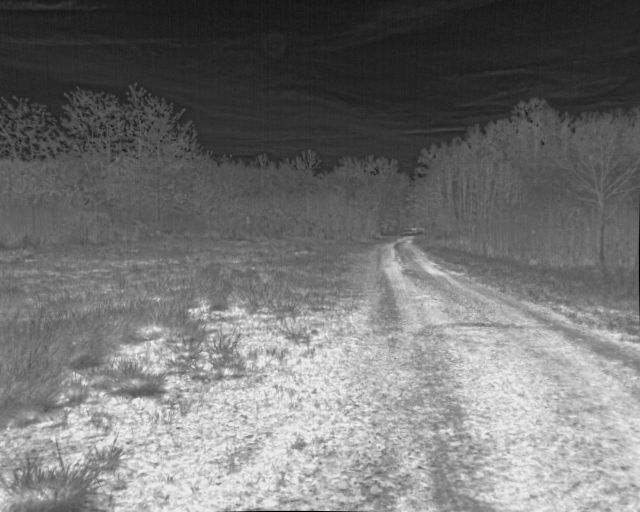}&
\includegraphics[width=1.7in,height=1in]{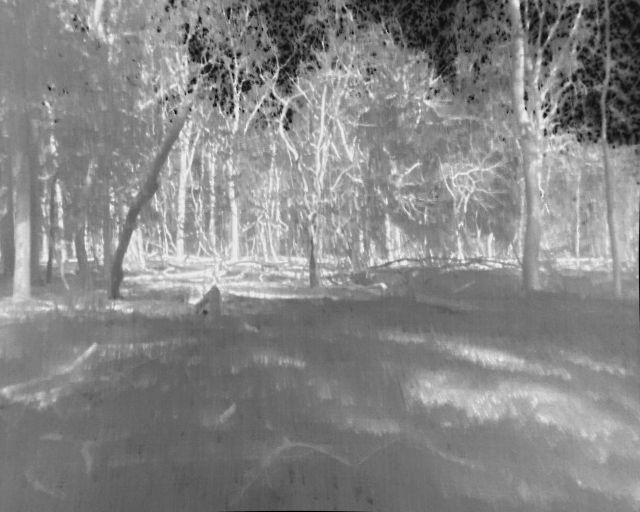}\\
(a) Forest Area      & (b) Trail\\
\end{tabular}
\caption{Example RGB images and corresponding semantic segmentation annotations and thermal data from the GO Dataset. Shown are (a) a trail scene and (b)a forest scene , along with their corresponding segmentation results and thermal images. Although captured at closely synchronized timestamps, the RGB and thermal views differ due to sensor placement and modality characteristics.} 
\label{fig:labels}
\end{figure}

%% file: figures/go_stat.tex
\begin{figure}[]
  \centering
  \includegraphics[width=\columnwidth]{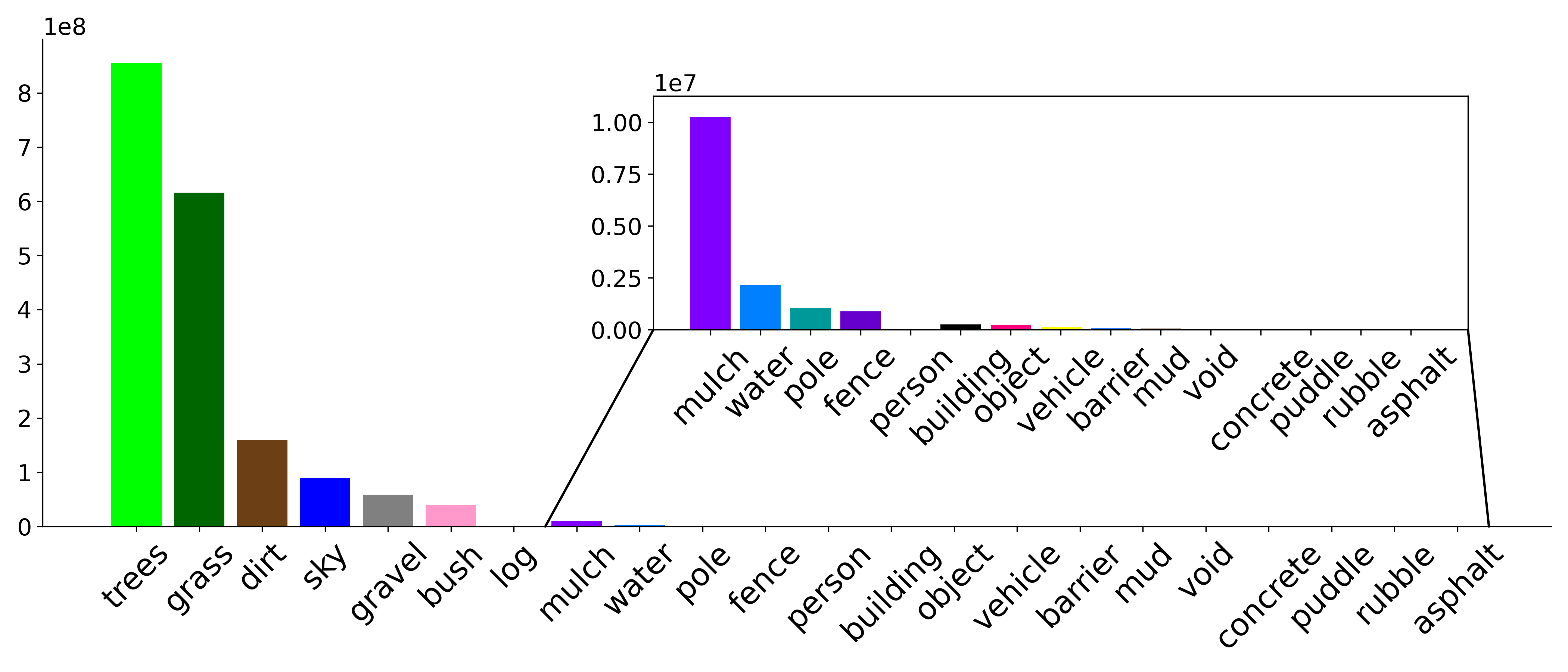}
  \caption{Image Label distribution. The tree, grass, dirt and sky constitute the major classes.}
  \label{fig:go_stat}
\end{figure}

%% file: sections/open_questions.tex
\section{Research Directions and Open Questions}
Although we only provide a preliminary evaluation of the dataset for a segmentation task, the utility of the GO dataset can be seen the following research areas:
\begin{itemize}
    \item \textbf{Robust Localization:}  The dataset provides accurate GPS ground truth trajectories synchronized with LiDAR, visual, radar, and NIR sensor data. This enables the development and evaluation of robust odometry, SLAM, and place recognition algorithms, particularly those leveraging novel modalities like radar \cite{overbyeRadarOnlyOffRoadLocal2023a} and NIR\cite{rankinPassivePerceptionSystem2005} or exploring advanced sensor fusion techniques for improved accuracy in off-road environments.

    \item \textbf{Advanced Perception:}  High-quality semantic annotations facilitate the development and evaluation of perception models for tasks such as semantic segmentation, object detection, and terrain classification. This rich labeled data enables the development of autonomous systems capable of robustly understanding and interpreting complex outdoor scenes.

    \item \textbf{Reliable Navigation:}  The dataset includes diverse routes with loops, varying speeds, and challenging terrains, providing a realistic testbed for evaluating and improving vision-based navigation models like ViNT \cite{shahViNTFoundationModel2023}. The multi-modal data enables these models to learn robust features and generalize effectively to complex off-road environments\cite{wangDriveAnywhereGeneralizable2024}. 
\end{itemize}
While the GO dataset offers a valuable resource for off-road robotics research, it also presents several challenges and open questions that warrant further investigation:
\begin{itemize}
    \item \textbf{Generalization to Unseen Environments}: How well do models trained on the GO dataset generalize to new and unseen off-road environments with different terrain types, vegetation, weather conditions, and object distributions?
    \item \textbf{Effective Multi-modal Fusion}: How can data from different sensor modalities (e.g., LiDAR, visual, radar, thermal, etc.) be effectively fused to achieve robust and accurate perception, localization, and navigation in challenging off-road settings?
    \item \textbf{Robustness to Degraded Features}: How can algorithms be made robust to degraded or missing sensor data caused by factors like dust, fog, rain, snow, or low-light conditions common in off-road environments?
    \item \textbf{Developing Standardized Evaluation Metrics}: What are the most appropriate metrics for evaluating the performance of different algorithms on the GO dataset, considering the specific challenges and requirements of off-road robotics tasks?
\end{itemize}

%% file: sections/conclusion.tex
\section{Conclusion}

This paper presents the Great Outdoors (GO) dataset, a comprehensive multimodal resource designed to advance robotics research in unstructured and off-road environments. The GO dataset integrates a diverse suite of sensors, high-quality semantic annotations for images, synchronized GPS/IMU trajectories, and data collected under challenging natural conditions. These characteristics make the GO dataset a valuable foundation for developing robust and adaptable autonomous systems capable of operating across diverse real-world outdoor scenarios.
Future extensions will focus on increasing environmental diversity across seasons, weather conditions, and times of day, while expanding data collection to more complex terrains beyond standard off-road trails. In addition, we plan to incorporate language annotations to support research on Vision-Language Models (VLMs) and Vision-Language-Action (VLA) frameworks in off-road contexts, enabling multimodal understanding and embodied reasoning in natural environments.

\section*{Acknowledgments}
Research reported in this paper was sponsored in part by the DEVCOM Army Research Laboratory under Cooperative Agreements W911NF-24-2-0023 (SARA CRA) and W911NF-23-2-0211.
The views and conclusions contained in this document are those of the authors and should not be interpreted as representing the official policies, either expressed or implied, of the DEVCOM Army Research Laboratory or the U.S. Government. The U.S. Government is authorized to reproduce and distribute reprints for Government purposes notwithstanding any copyright notation herein.